# On Sensitivity of the MAP Bayesian Network Structure to the Equivalent Sample Size Parameter

**Tomi Silander** and **Petri Kontkanen** and **Petri Myllymäki**
Complex Systems Computation Group (CoSCo)
Helsinki Institute for Information Technology (HIIT)
P.O.Box 68 (Department of Computer Science)
FIN-00014 University of Helsinki, Finland

## Abstract

BDeu marginal likelihood score is a popular model selection criterion for selecting a Bayesian network structure based on sample data. This non-informative scoring criterion assigns same score for network structures that encode same independence statements. However, before applying the BDeu score, one must determine a single parameter, the equivalent sample size $\alpha$. Unfortunately no generally accepted rule for determining the $\alpha$ parameter has been suggested. This is disturbing, since in this paper we show through a series of concrete experiments that the solution of the network structure optimization problem is highly sensitive to the chosen $\alpha$ parameter value. Based on these results, we are able to give explanations for how and why this phenomenon happens, and discuss ideas for solving this problem.

## 1 INTRODUCTION

BDeu (Bayesian Dirichlet equivalence uniform) score is a popular scoring metric for learning Bayesian network structures for complete discrete data. It corresponds to a set of plausible assumptions under which the posterior odds of different Bayesian network structures can be calculated, thus giving us the opportunity to find the maximum posterior (MAP) structure for the data. To obtain the BDeu score, we need a parameter called equivalent sample size $\alpha$ that expresses the strength of our prior belief in the uniformity of the conditional distributions of the network. A quick look at the Bayesian network learning literature reveals that there is no generally accepted "uninformative" value for the $\alpha$ parameter. In this paper we will show that this simple parameter often has a considerable effect on the MAP structure. While the result itself is not surprising, it is nonetheless bothersome, since $\alpha$ is not linked to the structure prior, still changing $\alpha$ changes the most probable structure.

We are not aware of any previous systematic empirical study on the role of $\alpha$. This may be due to the fact that Bayesian network structure learning is NP hard, and much of the effort has been put on heuristics that allow us to learn structures for datasets with realistic number of variables. However, recent advances in exact structure learning [1, 2] make it possible to study the role of $\alpha$ in structure learning systematically.

One may argue that learning the MAP structure is not a very desirable goal, and consequently, robustness problems of the MAP structure are not of big concern. One may adopt the engineering view on model selection [3] and insist that the model selection criterion should be related to the behavior of the model in some specific task such as joint probability prediction or classification. Especially, it may well be that different structures perform equally well in some specific task.

However, there is also a scientific view on the model selection [3] that calls for selecting the model that is most likely true. In this context the BDeu score is often used. It also coincides with the prequential model selection principle [4], which has a clear interpretation as predictive behavior. Furthermore, it has been shown that selecting the MAP model often performs well in prediction, while the fully Bayesian approach of averaging over all the structures is computationally infeasible due to the large number of structures. Finally, to our knowledge, the BDeu score is often used in practice. We want to emphasize that our intent is not to criticize or evaluate different scoring criteria, but to raise the awareness of the sensitivity of the MAP structure to the $\alpha$ parameter.

The paper consists of series of experiments, and it is structured as follows: In the next section we briefly review the BDeu score. In the following section we present the experiments and their results. We then try to give a more detailed account of how $\alpha$ affects the emergence of an arc in the MAP structure. We also discuss some ideas of circumventing this problem, and finally, we summarize the results and close with some conclusions.

## 2　BDeu SCORE

Bayesian networks are multivariate statistical models consisting of a qualitative component that encodes the conditional independence relations of the variables, and a quantitative component that determines the actual joint probability distribution of the variables. This paper deals with learning the qualitative part of the network, also called the structure of the network. The parameter under study, the equivalent sample size $\alpha$, is directly related to the prior probability of the parameters determining the joint probability distribution that satisfies the constraints imposed by the network structure. However, it turns out that this parameter also plays a big role in learning the network structure.

Observational data used for learning the structure is assumed to consist of $N$ data vectors each having $n$ nominal scale attributes (variables) $V = (V_1, \ldots, V_n)$, such that the $i^{th}$ variable $V_i$ has $r_i$ possible values that, without loss of generality, can be assumed to be $1, \ldots, r_i$. To explain the qualitative part of the network and the BDeu score [5, 6], we also need a method for enumerating all the possible value combinations (also called configurations) of any subset of variables. Details of this enumeration are not important: it is sufficient to uniquely identify the $j^{th}$ value combination of any variable subset $W, j \in \{1, \ldots, \prod r_i \mid V_i \in W\}$.

A network structure can be presented as a directed acyclic graph (DAG) containing $n$ nodes, one for each variable. The graph $G$ can be specified as a list of parent sets, $G = (G_1, \ldots, G_n)$, such that the set $G_i$ contains exactly those nodes from which there is an arc to the $i^{th}$ node. The structure of the network corresponds to a set of independence statements about variables. However, many different DAGs may correspond to the same set of independence assumptions. Therefore, it is natural to require that in the light of the data, all the different structures encoding the same set of independence assumptions should be equally probable. As shown by Heckermann et al. [5], this requirement of *likelihood equivalence* leads to very specific constraints on the quantitative part of the Bayesian network.

The quantitative part of a Bayesian network with structure $G$ consists of conditional multinomial probability distributions parameterized with vectors $\theta_{i|j}$ of length $r_i$, which give point probabilities of different values of $V_i$ in a situation where parents $G_i$ hold the $j^{th}$ value combination. When learning the Bayesian network from the data, we try to estimate these parameters. Assuming that these parameters are independent of each other, the only way to achieve likelihood equivalence is to model these parameters as if obeyed a Dirichlet distribution. More specifically, if a priori the $\theta_{i|j}$ is distributed according to $Dir(\alpha_1, \ldots, \alpha_{r_i})$, a posteriori $\theta_{i|j}$ is distributed as $Dir(\alpha_1 + N_{ij1}, \ldots, \alpha_{r_i} + N_{ijr_i})$, where $N_{ijk}$ indicates the number of data vectors in which $V_i$ has the value $k$ while its parents are in the $j^{th}$ configuration.

Relative sizes of the Dirichlet distribution hyperparameters $\alpha_*$ directly determine the expected values of the parameters $\theta_{i|j}$, so that having all the hyperparameters $\alpha_i$ equal corresponds to an expectation that different outcomes of $V_i$ are equally likely when its parents are in the $j^{th}$ configuration. In the case of equal $\alpha_i$ hyperparameters, their absolute values linearly correspond to the precision of the $\theta_{i|j}$, so that the larger the hyperparameters, the more persistently we believe in equiprobability of the outcomes of the $V_i$. When all hyperparameters equal 1.0, the Dirichlet distribution is actually a uniform distribution, i.e. all the possible vectors $\theta_{i|j}$ are equally likely. When all hyperparameters are less than one, we actually express disbelief of equiprobability of $\theta_{i|j}$ favoring the parameter vectors in which some coordinates are big and others small.

In this paper we concentrate on a case in which we have no a priori reason to judge any network structure or any value combination of a variable set more probable than others. We also want our assessment of the quality of the Bayesian network structure to adhere to the requirement of likelihood equivalence.

It can be shown [5] that the requirement of likelihood equivalence necessarily leads to setting the hyperparameters of the a priori Dirichlet distribution of $\theta_{i|j}$ using the method of equivalent samples. In this method, we imagine observing a certain number $\alpha$, called the *equivalent sample size*, of data vectors. Since we further aim at being non-informative about $\theta_{i|j}$ a priori, we must we have observed equally many vectors of all different kinds. Note that this method often stretches our imagination, since it may imply observing a certain data vector a fractional number of times (for example, we imagine seeing all the possible data vectors 0.02 times.") After picking the value for $\alpha$, the hyperparameters for $\theta_{i|j}$ are set to $\frac{\alpha}{r_i q_i}$, where $q_i$ is the number of possible value combinations of $V_i$'s parents $G_i$ (or 1 if $V_i$ has no parents, i.e., $G_i = \emptyset$). This gives us the following formula for the marginal likelihood of the network $P(D|G, \alpha)$:

$$\prod_{i=1}^{n} \prod_{j=1}^{q_i} \frac{\Gamma(\frac{\alpha}{q_i})}{\Gamma(\frac{\alpha}{q_i} + \sum_{k=1}^{r_i} N_{ijk})} \prod_{k=1}^{r_i} \frac{\Gamma(\frac{\alpha}{q_i r_i} + N_{ijk})}{\Gamma(\frac{\alpha}{q_i r_i})}. \quad (1)$$

This likelihood, the BDeu score, was proposed early on by Buntine [6], and in [5] authors note that using wrong informative priors is dangerous and BDeu should be preferred. The actual numerical value of the marginal likelihood can be very small, so it is customary to use its logarithm as a score. Taking the logarithm is straightforward by turning the products into sums, ratios into subtractions and using $\log \Gamma$ (available in many numerical libraries) instead of $\Gamma$.

The learning task is to find the network structure that maximizes the BDeu (or equivalently, its logarithm). Assuming uniform priors for network structures (or the equivalence classes of structures that encode the same independence assumptions), maximizing the BDeu score equals maximizing the posterior probabilities of the structures.

## 3 EXPERIMENTS

There has been some recent development in finding the globally optimal Bayesian network structures using decomposable scores such as BDeu. In order to study the effect of $\alpha$ on the MAP structure, we used the bene-software that is freely available at http://b-course.hiit.fi/bene, and 20 different UCI data sets [7] (Table 1). Some of the data sets contain continuous values and missing values, so the sets have to be discretized and imputed before learning. In all cases, the continuous variables were discretized to three equal width bins, and the imputation was made by randomly selecting a value to be imputed according to the empirical distribution of the (possibly discretized) variable.

### 3.1 HOW MUCH CAN THE MAP STRUCTURE VARY

In [8] Steck and Jaakkola showed that asymptotically, as $\alpha$ goes to zero, the addition or deletion of an arc in a Bayesian network is infinitely favored or disfavored, and that the preference depends on effective degrees of freedom, a measure that is defined in terms of sufficient statistics $N_{ijk}$ that equal zero. They also suggest that in the other extreme, when $\alpha$ approaches infinity, the number of arcs in the MAP structure most probably increases. These results are significant, but since they are asymptotic they may not sound alarming enough. Steck and Jaakkola also conducted a small experiment that shows slight variation in the structure when $\alpha$ varies in the range $1,\ldots,1000$. Our study will show that the variation in MAP structure cannot be brushed aside by referring to asymptotics: there is considerable variation even with reasonable values of $\alpha$.

We first wanted to empirically verify the findings of Steck and Jaakkola. As an initial experiment, we learned the best network for the Yeast data (9 variables, 1484 data vectors) with different values of the equivalent sample size $\alpha$. By going from very small values of $\alpha$ (2e-20) to very large values of $\alpha$ (34000), it was possible to get any number of arcs, i.e. from 0 to 36, to the MAP network structure (Figure 1).

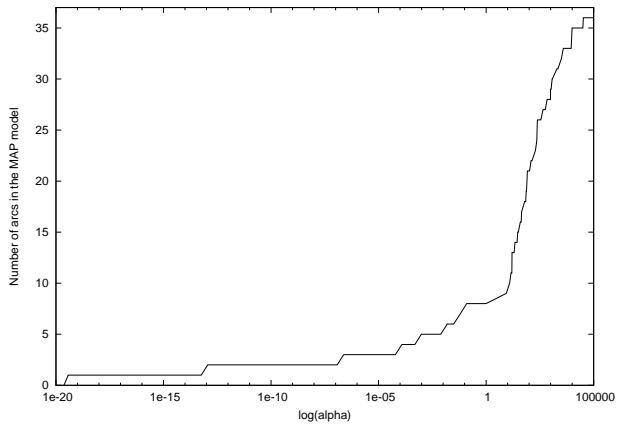

Figure 1: Number of arcs in the BDeu optimal network for the Yeast data as a function of $\alpha$.

This immediately prompted a question if this is possible for other data sets too. Therefore, we attempted to do the same for 20 different data sets using $\alpha$ between 1e-20 and 1 000 000. The results are summarized in Table 1, in which each of the 20 data sets is listed together with its number of vectors (N), number of variables (n), and the average number of values per variable (vals). Other columns of the table will be explained later.

The results of our pursuit for different arc counts are summarized in column #arcs that show how many different arc counts (out of maximum possible) were obtained using different $\alpha$'s. The range of arc counts (i.e., the smallest and highest number of arcs obtained by varying $\alpha$) is listed in column "range". Finally, the range of $\alpha$ values needed to produce the observed variation is listed in column $\alpha$-range.

The results show that we cannot always produce any number of arcs in our MAP model by just tweaking $\alpha$. However, for many data sets it is possible to get a considerable amount of different arc counts. In any case, these results prompt further questions about the nature of the phenomenon. First of all, some of these effects are produced using unrealistic values of $\alpha$. Could it be that reasonable values of $\alpha$ are "safe". The arc count of the MAP model tends to increase with $\alpha$, but is this always the case?

Table 1: Summary of the results of experiments.

| Data | N | n | vals | $\alpha$-range | #arcs | range | $\alpha^I$ | $\alpha^*$ | $\#_1^{10}$ | $\#_1^{100}$ |
|---|---|---|---|---|---|---|---|---|---|---|
| balance | 625 | 5 | 4.6 | [0.03 , 0.04] | 2/11 | 0–4 | 1…100 | 48 | 1 | 1 |
| iris | 150 | 5 | 3.0 | [5e-14, 122] | 9/11 | 2–10 | 1…3 | 2 | 4 | 7 |
| thyroid | 215 | 6 | 3.0 | [1e-6 , 488 ] | 12/16 | 4–15 | 2…2 | 2 | 5 | 10 |
| liver | 345 | 7 | 2.9 | [4e-7 , 4e+4] | 22/22 | 0–21 | 3…6 | 4 | 4 | 12 |
| ecoli | 336 | 8 | 3.4 | [3e-20, 8e+3] | 25/29 | 3–28 | 7…10 | 8 | 2 | 12 |
| abalone | 4177 | 9 | 3.0 | [6e-18, 4e+4] | 26/37 | 10–36 | 6…6 | 6 | 6 | 12 |
| diabetes | 768 | 9 | 2.9 | [2e-13, 4e+4] | 35/37 | 0–36 | 3…5 | 4 | 5 | 16 |
| post op | 90 | 9 | 2.9 | [2.29 , 7e+4] | 29/37 | 0–31 | 3…5 | 3 | 6 | 16 |
| yeast | 1484 | 9 | 3.7 | [2e-20, 4e+4] | 37/37 | 0–36 | 1…6 | 6 | 2 | 17 |
| cancer | 286 | 10 | 4.3 | [5e-14, 3e+5] | 38/46 | 0–39 | 6…10 | 8 | 3 | 15 |
| shuttle | 58000 | 10 | 3.0 | [2e-19, 4e+3] | 24/46 | 20–45 | 1…3 | 3 | 5 | 13 |
| tictac | 958 | 10 | 2.9 | [4e-17, 4e+3] | 20/46 | 8–33 | 51…62 | 51 | 5 | 24 |
| bc wisc | 699 | 11 | 2.9 | [2e-8 , 8e+4] | 48/56 | 8–55 | 7…15 | 8 | 7 | 24 |
| glass | 214 | 11 | 3.3 | [1e-11, 2e+5] | 39/56 | 8–53 | 5…6 | 6 | 8 | 21 |
| page | 5473 | 11 | 3.2 | [0.125, 2e+5] | 41/56 | 15–55 | 3…3 | 3 | 8 | 20 |
| heart cl | 303 | 14 | 3.1 | [3e-5 , 1e+6] | 80/92 | 0–86 | 13…16 | 13 | 4 | 20 |
| heart hu | 294 | 14 | 2.6 | [9e-10, 5e+5] | 67/92 | 3–75 | 5…6 | 5 | 6 | 30 |
| heart st | 270 | 14 | 2.9 | [13.4 , 8e+5] | 71/92 | 19–85 | 7…10 | 10 | 7 | 22 |
| wine | 178 | 14 | 3.0 | [1e-4 , 5e+5] | 60/92 | 8–85 | 8…8 | 8 | 8 | 20 |
| adult | 32561 | 15 | 7.9 | [1e-20, 3e+5] | 26/106 | 15–79 | 48…58 | 50 | 5 | 17 |

## 3.2 NATURE OF VARIATION

Larger values of $\alpha$ tend to produce MAP structures with more arcs. However, it appears that this rule also has some exceptions. Sometimes, like with the Post Operative data with $\alpha$ values 50 and 55, the larger $\alpha$ actually produces less arcs (14 instead of 15). It is also possible that increasing $\alpha$ changes the MAP model structure even if the actual number of arcs stays the same. This happens with the Thyroid data when we change $\alpha$ from 2 to 3. In this case the skeleton of the network stays the same, but one V-structure changes the place yielding a different independence model. Had the relation been so simple that increasing $\alpha$ always added arcs to previous MAP models, we could get a simple picture of robustness of a MAP model by trying smaller and bigger values of $\alpha$. While we can still do that, and it is indeed advisable, the results are not necessarily easy to interpret because of the possibly non-monotonic variation in arc counts. In practice, trying systematically very many different values of $\alpha$ is usually not feasible, since even learning a MAP model with one single $\alpha$ is already very hard.

## 3.3 VARIATION WITH REALISTIC VALUES OF $\alpha$

With the Liver data, $\alpha$ values of 0.01, 0.05, 0.10, 1.00, 2.00, 3.00, 9.00, and 13.00 all yield different MAP models with different number of arcs (1,…,8).

The log BDeu scores of these eight models are shown in Figure 2 as a function of the $\alpha$ in a range [1.0, 6.0]. Notice that the difference of 5 in the BDeu score corresponds to the marginal likelihood ratio of about 150.

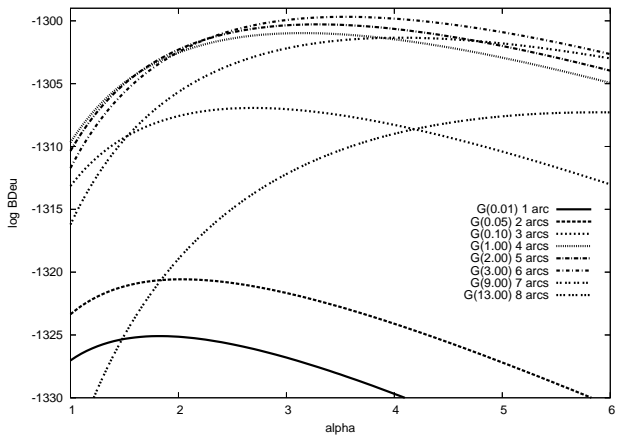

Figure 2: BDeu scores of different MAP models for the Liver data as a function of $\alpha$.

We can clearly conclude that MAP structures may vary also within realistic values of $\alpha$. In the Breast Cancer Wisconsin data, $\alpha$ values 1.00, 1.02, and 1.04 yield different MAP structures with different numbers of arcs (12, 13, and 14 respectively), see Figure 3. Moreover, this may also happen when the sample size is relatively large. In the Adult data set, $N = 32561$, $\alpha$ values 2, 4, 8, and 16 all produce different MAP models with arc counts 28, 29, 31 and 33 respectively.

For all data sets, we also tested $\alpha$-values 1,2,...,100. Columns $\#_1^{10}$ and $\#_1^{100}$ of Table 1 indicate how many different MAP models (i.e., network structures that encode different independence assumptions) were found with $\alpha$ values 1,...,10 and 1,...,100 respectively.

## 4 THE REASON FOR AN ARC

A simple answer why a certain arc appears in the MAP model is that without that arc the graph gets an inferior BDeu score. While not controversial, an explanation like this is hardly enough for a scientist that misses his favorite arc.

In order to get a better understanding of the behavior of the BDeu score, and to see how changing $\alpha$ just slightly can make the difference, we delve into the details of the BDeu score using the example (Figure 3) mentioned in the previous section.

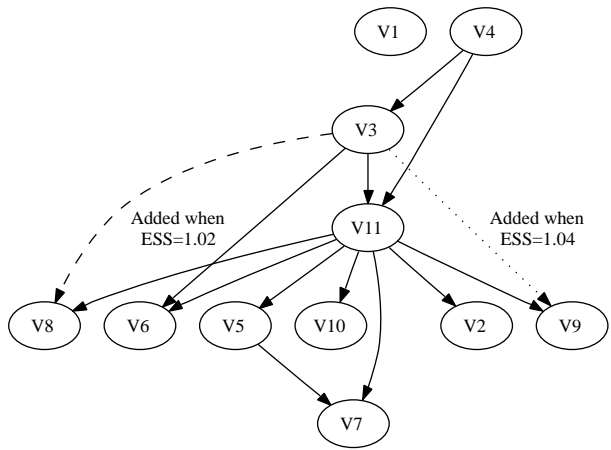

Figure 3: The MAP models for the BC Wisconsin data with $\alpha=1.0$, $\alpha=1.02$, and $\alpha=1.04$

We start our study by regrouping the factors of the BDeu score formula:

$$\prod_{i=1}^{n}\prod_{j=1}^{q_i} \frac{\Gamma(\sum_{k=1}^{r_i} \frac{\alpha}{q_i r_i})}{\prod_{k=1}^{r_i} \Gamma(\frac{\alpha}{q_i r_i})} \frac{\prod_{k=1}^{r_i} \Gamma(N_{ijk}+\frac{\alpha}{q_i r_i})}{\Gamma(\sum_{k=1}^{r_i} N_{ijk}+\frac{\alpha}{q_i r_i})}. \quad (2)$$

By using the multinomial Beta functions

$$B(\alpha_1,\ldots,\alpha_K) = \frac{\prod_{i=1}^{K}\Gamma(\alpha_i)}{\Gamma\left(\sum_{i=1}^{K}\alpha_i\right)}, \quad (3)$$

we can write the score as

$$\prod_{i=1}^{n}\prod_{j=1}^{q_i} \frac{B(N_{ij1}+\frac{\alpha}{q_i r_i},\ldots,N_{ijr_i}+\frac{\alpha}{q_i r_i})}{B(\frac{\alpha}{q_i r_i},\ldots,\frac{\alpha}{q_i r_i})}. \quad (4)$$

We will first study the denominator of the formula above. The denominator does not depend on the data and, as we will shortly see, it acts as a complexity penalizing factor. However, the magnitude of this penalty depends on $\alpha$. To see this, let us consider assigning another variable with $K$ values as a parent of the $i^{th}$ variable that already has $q_i$ parent configurations. The change of the denominator term in the logarithmic version of the BDeu score would be $K \times \log B(\frac{\alpha}{Kq_i r_i},\ldots,\frac{\alpha}{Kq_i r_i}) - \log B(\frac{\alpha}{q_i r_i},\ldots,\frac{\alpha}{q_i r_i})$, which is depicted in Figure 4 as a function of $\frac{\alpha}{q_i r_i}$ for different K.

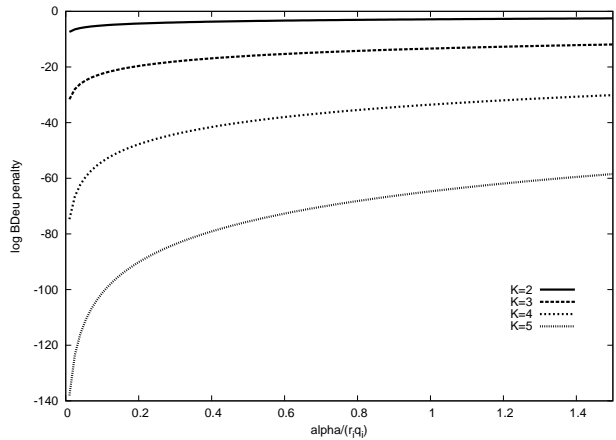

Figure 4: Change in the denominator log BDeu after adding another parent with K values to the $i^{th}$ variable, as a function of $\frac{\alpha}{q_i r_i}$.

Notice that there are $q_i$ of these penalties in the score. As we readily observe, the larger the K the greater the penalty of adding the arc, so the increase in the number of model parameters is penalized by this term. We also notice that increasing the $\alpha$ actually makes the penalty smaller.

In our example, the variable V8 ($r_8 = 3$) already has a parent V11 ($r_{11} = 2$), thus $q_8 = 2$. The new parent candidate, V3, has three different values ($K = r_3 = 3$), thus our penalty for drawing a new arc is $2 \times -16.02$ when $\alpha=1.0$ and $2 \times -15.94$ when $\alpha=1.02$.

The crux of the score lies in the numerator that depends on data through sufficient statistics $N_{ijk}$ and the fractions of $\alpha$. In general, adding a parent with K values splits its child's frequency histograms defined by $N_{ij*}$ into K smaller histograms. Only if these histograms are more informative than the original ones, the addition may be justified by the score.

Log multinomial Beta favors small informative histograms, which can be seen in Figure 5 in which the log Beta is plotted for histograms ($M \times (N_1, N_2, N_3)$) of different entropies and multipliers M. (For example, $3 \times (5, 1, 0) = (15, 3, 0)$.) From the approximate linearity of the graphs we may conclude that splitting a histogram evenly into two histograms is slightly favored by the log Beta function. However, the denominator penalizes the division so that the net effect is negative.

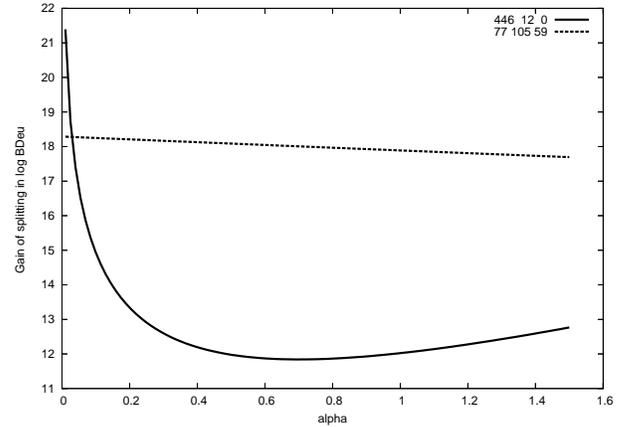

Figure 6: Changes in terms of BDeu score numerator caused by adding an arc from V3 to V8.

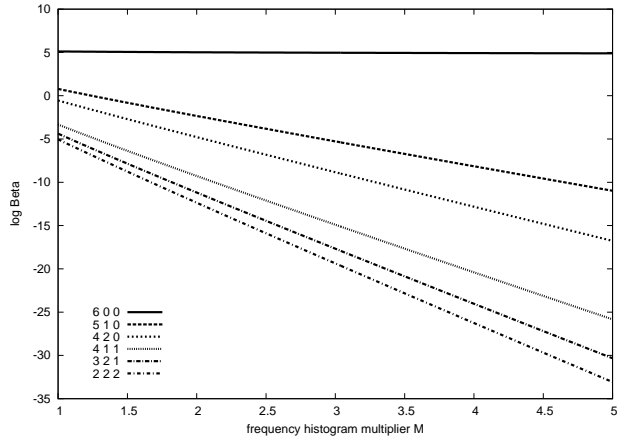

Figure 5: Log Beta for different histograms and their multiples.

In our example, before the arc from V3 to V8 is added, the sufficient statistics for V8 can be expressed by two histograms (one for each value of the parent V11): (446, 12, 0) and (77, 105, 59). Adding an arc from V3 to V8 splits both of these histograms in three. The first one will be split into (442, 11, 0), (2, 1, 0), and (2, 0, 0), and the second one will be split into (34, 29, 5), (22, 39, 12), and (21, 37, 42). The actual arguments for the beta function in the numerator are these frequencies plus the $\alpha$ terms which before the split are $\alpha/2$ and after splitting $\alpha/6$. The positive change in the log BDeu numerator caused by these splits (i.e. adding an arc) is presented in Figure 6 as a function of $\alpha/2$.

We are now ready to calculate the effect of adding an arc for $\alpha=1.00$ and $\alpha=1.02$. In both cases the calculation can be expressed as a sum of three factors: two times the penalty for splitting histograms in three, gain of splitting the first histogram, and the gain in splitting the second histogram. For $\alpha=1.00$ the numbers are $2 \times -16.02 + 13.73 + 18.22 = -0.09$, while for $\alpha=1.02$ we get $2 \times -15.94 + 13.68 + 18.22 = 0.02$. We may conclude that increasing $\alpha$ from 1.00 to 1.02 eased the penalty for adding the arc just enough to justify it.

The gain from splitting the first histogram seems to behave oddly while the other term behaves nicely. This is due to the zero sufficient statistics $N_{ijk}$ that make the numerator of the Beta function contain Gamma functions that depend on the fractions of the $\alpha$ only. This observation is in line with the findings of Steck and Jaakkola which emphasize the role of zero sufficient statistics in the asymptotic behavior of the MAP model selection as a function of decreasing $\alpha$.

## 5  THE "BEST MAP MODEL"

### 5.1  INTEGRATING OUT $\alpha$

For a Bayesian it is natural that different priors lead to different MAP models. However, it is annoying that a parameter prior $\alpha$ intended to convey ignorance may make a big difference when selecting the network structure. Faced with the uncertainty of selecting a value for $\alpha$, a Bayesian way would be to integrate $\alpha$ out, and then choose the most probable structure. However, learning Bayesian network structures is NP-hard and also computationally very demanding; for large data sets one must adhere to heuristic search methods. All this makes integrating out $\alpha$ currently impossible in practice for data sets with many variables (say, 30 and above).

In this paper we use smaller data sets to study which of the MAP models are selected if we have the luxury of integrating out $\alpha$. The idea is to find out which $\alpha$ value yields the MAP model that is also selected by integrating $\alpha$ out. To do this, we have to specify a prior distribution $P(\alpha|G)$ for $\alpha$. While it is not obvious how to do that, it is hardly much more demanding than picking just one $\alpha$. It is also unclear how the prior for $\alpha$ should depend on the model structure $G$. For the sake of simplicity we might say it does not depend on $G$ at all. Furthermore, we might take a uniform prior for structures, and all this would lead to a $\alpha$-prior weighted BDeu score: pick the model that "on average for different values of $\alpha$" gives a high BDeu score, the average being an $\alpha$ prior weighted average:

$$\begin{aligned} P(G|D) &= \int_\alpha P(G,\alpha|D) \quad (5) \\ &= \frac{P(G)}{P(D)} \int_\alpha P(D|G,\alpha)P(\alpha|G). \end{aligned}$$

However, we know no closed-form solution for this integral, and numerical integration is costly. To get an idea of what kind of structures are selected by integrating out $\alpha$, we assigned a uniform prior to $\alpha$ in the range from 1 to 100. We then selected all the different MAP models yielded by the $\alpha$ values in that range and studied the posterior distribution of these models.

In Figure 7 one can find posterior probabilities of 10 of the 22 network structures for the Glass data that can be found for $\alpha$ values 1—100. We see that the the MAP structure for $\alpha = 5$ ($\alpha = 6$ yields the same MAP model) gets the highest posterior probability when we integrate out $\alpha$.

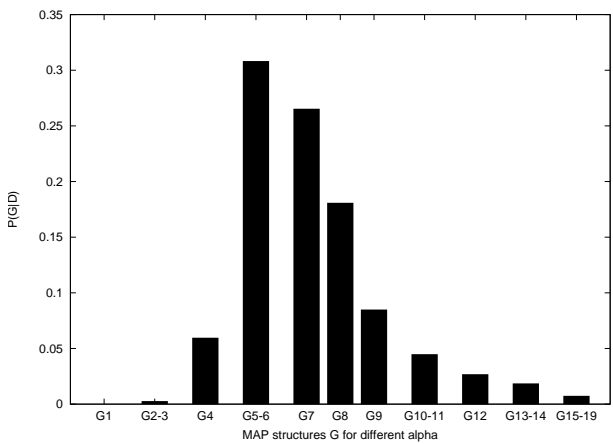

Figure 7: Posterior distribution of MAP models for the Glass data.

For other data sets, the $\alpha$ values selecting the same MAP models as the integration method does are listed in the Table 1 as a column $\alpha^I$. We notice that integrating out $\alpha$ yields models that are MAP models for reasonable (i.e. between 1 and 100) values of $\alpha$.

For non-orthodox Bayesians, there is also an option of selecting $\alpha$ and its MAP model $G^*$ that maximizes the marginal likelihood:

$$(G^*, \alpha^*) = \underset{(\alpha, G)}{\arg\max} \, P(D|G, \alpha). \quad (6)$$

This can be justified by a prequential model selection principle. The marginal likelihood can be decomposed by the chain rule as:

$$P(D|G,\alpha) = \prod_{j=1}^N P(d_j|\{d_1,\ldots,d_{j-1}\},G,\alpha), \quad (7)$$

which can be seen as a sequential prediction task. The prequential model selection principle says that the winning $(G,\alpha)$-pair is worth selecting.

The column $\alpha^*$ in Table 1 indicates the prequential score maximizing $\alpha$. For our data sets, the prequential score maximizing $\alpha$s range from 2 (for the Iris-data) to 51 (for the Tic-tac-toe data). We immediately notice that good values of $\alpha$ lie in the reasonable range: no $\alpha$ is very small or larger than number data vectors. However, a convenient choice of $\alpha=1$ appears to be slightly too small. One might safely conclude that this method does not readily point to any easy heuristic of selecting the $\alpha$. In practice, when "forced" to pick a value of $\alpha$, selecting the $\alpha$ to be the average number of values per variable could be a possible strategy.

To our surprise, in our experiments, the selected $\alpha^*$ always yields the same model as the integration method (since for all data sets, $\alpha^* \in \alpha^I$). This indicates that the volume around the maximum (marginal) likelihood $\alpha$ is large enough to dominate the integral.

It is also worth mentioning that maximizing $\alpha$ can be seen as an extreme case of integrating out $\alpha$ for different prior distributions in which a single value of $\alpha$ has probability 1. This will naturally generalize to studying of network structure selection under different prior distributions: instead of asking what is the correct $\alpha$, we may ask what is the correct (or non-informative) prior distribution for $\alpha$. Since an extreme prior clearly affects the MAP-structure, one might predict that other very informative priors will too.

## 6 DISCUSSION

We have shown that under the BDeu score, the MAP structure can be very sensitive to the parameter prior $\alpha$. In a way, this emphasizes the need to be Bayesian about the structure too [9]. However, since learning a Bayesian network structure is computationally a resource consuming task, we often have no other choice but to learn one single structure, the most probable one, so the question about the best parameter prior remains. Naturally, after learning the structure $G$ with one $\alpha$, it is possible to evaluate the BDeu score for structures close to the $G$ (say, one arc deletion, addition or reversal from the $G$) and for different values of $\alpha$. Our result emphasize the need for such a check. The results also show that examining neighboring structures and $\alpha$ values will probably lead to discovering different MAP-structures, so we face again the question of selecting the model. As much as we love clear answers, it is better to be aware of the brittleness of our model selection than to simply make convenient assumptions (say $\alpha = 1.0$) and never look back.

We have studied two ways to determine the "correct" $\alpha$. The results do not give a definite answer, and from the Bayesian point of view, the whole question may appear heretic or a non-problem: the prior should not be tuned after seeing the data and the result of the analysis should be the posterior of the structures instead of a single MAP model. While philosophically sound, these views are hard to implement fully in practice. In limited scale both maximizing the $\alpha$ and integrating it out are possible after we limit the number of candidate models, which can be done after heuristically learning the structure. Integrating out $\alpha$ raises the question about its prior distribution and its effect in model selection. This question is a subject for further study, as is the question which one of these two methods, if and when they differ, yields better predictions.

There are other common decomposable score equivalent criteria such as BIC and AIC that do not need any parameters. However, these scores are not without their problems since they are derived as asymptotic approximations. The information-theoretic normalized maximum likelihood (NML) approach [10] offers an interesting, alternative perspective to this problem: the NML does not require an explicitly defined parameter prior, but still offers an exact non-informative model selection criterion closely related to marginal likelihood with the Jeffreys prior [10]. Additionally, the NML is score equivalent like BDeu. Unfortunately computationally efficient algorithms for calculating the NML have been found only for certain restricted Bayesian network structures [11], but the situation will hopefully change in the future.

**Acknowledgements**

This work was supported in part by the Finnish Funding Agency for Technology and Innovation under projects PMMA, KUKOT and SIB, by the Academy of Finland under project CIVI, and by the IST Programme of the European Community, under the PASCAL Network of Excellence, IST-2002-506778. This publication only reflects the authors' views.

## References

[1] T. Silander and P. Myllymäki. A simple approach for finding the globally optimal Bayesian network structure. In *Proceedings of the 22nd Conference on Uncertainty in Artificial Intelligence (UAI-2006)*, pages 445–452, 2006.

[2] M. Koivisto and K. Sood. Exact Bayesian structure discovery in Bayesian networks. *Journal of Machine Learning Research*, 5:549–573, May 2004.

[3] D. Heckerman and D. Chickering. A comparison of scientific and engineering criteria for bayesian model selection. In *Proceedings of the Sixth International Workshop on Artificial Intelligence and Statistics*, pages 275–281, Ft. Lauderdale, Florida, January 1997.

[4] A.P. Dawid. Statistical theory: The prequential approach. *Journal of the Royal Statistical Society A*, 147:278–292, 1984.

[5] D. Heckerman, D. Geiger, and D.M. Chickering. Learning Bayesian networks: The combination of knowledge and statistical data. *Machine Learning*, 20(3):197–243, September 1995.

[6] W. Buntine. Theory refinement on Bayesian networks. In B. D'Ambrosio, P. Smets, and P. Bonissone, editors, *Proceedings of the Seventh Conference on Uncertainty in Artificial Intelligence*, pages 52–60. Morgan Kaufmann Publishers, 1991.

[7] S. Hettich and D. Bay. The UCI KDD archive, 1999. http://kdd.ics.uci.edu/.

[8] Harald Steck and Tommi Jaakkola. On the dirichlet prior and bayesian regularization. In Suzanna Becker, Sebastian Thrun, and Klaus Obermayer, editors, *NIPS*, pages 697–704. MIT Press, 2002.

[9] N. Friedman and D. Koller. Being bayesian about network structure: A bayesian approach tostructure discovery in bayesian networks. *Machine Learning*, 50:95–126, 2003.

[10] J. Rissanen. Fisher information and stochastic complexity. *IEEE Transactions on Information Theory*, 42(1):40–47, January 1996.

[11] P. Kontkanen, P. Myllymäki, W. Buntine, J. Rissanen, and H. Tirri. An MDL framework for data clustering. In P. Grünwald, I.J. Myung, and M. Pitt, editors, *Advances in Minimum Description Length: Theory and Applications*. The MIT Press, 2006.